\documentclass[12pt]{article}

\usepackage[margin=1in,letterpaper]{geometry}

\usepackage{amsmath}
\usepackage{amssymb}
\usepackage{graphicx}

\usepackage{booktabs}
\usepackage{stfloats}
\usepackage[table,xcdraw]{xcolor}
\usepackage{authblk}
\definecolor{lightblue}{rgb}{0.678, 0.847, 0.902}
\definecolor{lightyellow}{rgb}{1.0, 1.0, 0.878}
\definecolor{blue}{RGB}{173, 216, 230}
\definecolor{yellow}{RGB}{255, 255, 153}

\usepackage[none]{hyphenat}

\begin{document}
	
		
		
		
		\title{Meta-Learning for Classifier Selection in Image Datasets: A Feature-Driven Framework for Accuracy Prediction}
		
		
		\author[1]{Zahra Nabizadeh Shahre Babak} 
		\author[2]{Farzaneh Koohestani} 
		\author[2]{Nader Karimi} 
		\author[1]{Shahram Shirani} 
		\author[3]{Shadrokh Samavi} 
		\affil[1]{\small Department of Electrical and Computer Engineering, McMaster University, Hamilton L8S 4L8, Canada}
		\affil[2]{\small Department of Electrical and Computer Engineering, Isfahan University of Technology, Isfahan 8415683111, Iran}
			
		\affil[3]{\small Department of Computer Science, Seattle University, Seattle, WA 98122-1090, USA}
			
		\date{}
		\maketitle
		\begin{abstract}
			No Free Lunch theorem implies that any performance gains achieved by a classifier on a particular image distribution are necessarily offset by a loss of performance over the set of all possible problems; thus, no single model is universally optimal. Selecting the most suitable classifier for image datasets is a critical yet challenging task due to the intrinsic complexity and diversity of images. This paper proposes a meta-learning framework that leverages a comprehensive set of meta-features capturing dataset complexity to predict classifier performance without exhaustive training. By extracting and selecting features using methods such as autoencoders, pre-trained networks, and dimensionality reduction techniques, we train regression models to efficiently estimate classifier accuracies. Additionally, clustering techniques are employed to group classifiers with similar performance patterns, simplifying the recommendation process. The datasets used span a wide range of concepts including nature, animals, numbers, motorcycles, medical images, and human bodies to ensure broad generalization. Evaluated on 56 diverse image datasets, our approach achieves an average ranking prediction accuracy exceeding 86\%, demonstrating its effectiveness in guiding model selection. This scalable and interpretable framework provides a practical solution to improve classification performance while reducing computational costs.
			
			\vspace{0.5em}
		\end{abstract}
		
		
	
		
		
		\section{Introduction}
		\label{sec1}
		
		In recent years, machine learning algorithms have gained significant popularity and are now widely used across various applications. These algorithms are designed to learn from the distribution of input data to perform diverse tasks effectively. One key factor influencing the learning process is meta-learning (MtL), which involves learning how to learn. As a result, substantial research has been conducted in this area, leading to the development of various techniques aimed at improving the performance of Machine Learning (ML) algorithms.
		
		In the context of ML, particularly with image datasets, a comprehensive understanding of the data is essential for successful classification. While numerous studies have focused on measuring dataset complexity, few specifically address the complexity of image datasets in classification tasks. Understanding dataset complexity is crucial for selecting the most suitable models, as it guides decisions related to preprocessing, feature extraction, and model selection. Many decision-making challenges in machine learning can be formulated as ML problems, especially if a suitable dataset can be generated for the decision-making process itself. The following section discusses articles that explore the use of dataset complexity for decision-making purposes.
		\subsection{Complexity Measure}
		\label{subsec1}
		
		To measure the complexity of datasets, various perspectives can be considered. Each perspective highlights a specific property of the datasets that can be treated as meta-features. Research on meta-features often involves proposing new feature categories or introducing novel meta-features to enhance the learning process. For instance, in \cite{lorena2019complex}, the authors classified meta-features related to dataset characterization and classification complexity into five categories: feature-based, linearity, neighborhood, network, dimensionality, and class imbalance. In \cite{pimentel2019new}, Pimentela and Carvalho introduced a new set of meta-features that outperformed previous techniques in algorithm recommendation tasks. They extracted three sets of features—statistical, distance-based, and evaluation measures—from datasets and used them to recommend the best clustering algorithm. The paper \cite{rivolli1808characterizing} systematizes data characterization measures for classification datasets and introduces the Meta-Feature Extractor (MFE) to enhance reproducibility in meta-learning research.
		
		There are various meta-features that describe datasets and can provide insights into their complexity. Some of these meta-features are applicable to image or 2D datasets, while others are specific to 1D datasets. These meta-features offer different perspectives on the dataset’s characteristics. Below, we will explain these meta-features based on their respective approaches. 
		
		Some meta-features describe datasets using network-based approaches. By mapping each dataset into a graph or tree structure, these features capture dataset characteristics. In \cite{perez2016graph}, the authors introduced a network metric called closeness centrality, which measures the relative distance of a node to all others based on shortest path. Another widely used feature, degree centrality \cite{bang2008digraphs}, quantifies the number of connections a node has, indicating its importance. Betweenness centrality \cite{freeman1977set} measures a node's significance by calculating the number of shortest paths passing through it. A related measure, edge betweenness centrality \cite{brandes2008variants}, evaluates the importance of edges based on shortest paths. PageRank \cite{page1999pagerank} ranks nodes solely by the structure of their connections, while eigenvector centrality \cite{newman2008mathematics} considers both the number of connections and the significance of those connections. A spectral clustering-based metric partitions a graph into subgraphs by minimizing inter-cluster edge weights. This feature maps samples into a transformed space where similar images are clustered, followed by the computation of the Laplacian matrix and the Cumulative Spectral Gradient (CSG) using its eigenvalues and eigenvectors \cite{branchaud2019spectral}. 
		
		Several meta-features evaluate clustering output. The Silhouette Coefficient \cite{rousseeuw1987silhouettes} assesses how well each data point fits within its assigned cluster relative to other clusters. The Calinski-Harabasz (CH) Index \cite{calinski1974dendrite} measures clustering quality based on the ratio of between-cluster dispersion to within-cluster dispersion, where higher values indicate better clustering. The Davies-Bouldin (DB) Index \cite{davies1979cluster} compares the average within-cluster distances to between-cluster distances, with lower values signifying better separation. Similarly, the Dunn Index (DI) \cite{dunn1974well} calculates the ratio of minimum inter-cluster distance to maximum intra-cluster distance. Another clustering-related feature, normalized relative entropy, also known as Kullback-Leibler divergence, quantifies dissimilarity between cluster probability distributions \cite{pimentel2019new}. The Int Index \cite{bezdek1998some} evaluates cluster compactness and separation based on intra- and inter-cluster distances.
		
		Entropy-based features describe dataset complexity. Shannon entropy, GLCM entropy, and delentropy were used in \cite{rahane2020measures} to rank image datasets based on complexity and correlate their rank with deep learning performance. Other information-theoretic features include Mutual Information (MI), which quantifies dependency between sample attributes and target labels, and the noisy feature ratio, which measures the proportion of irrelevant attributes \cite{rivolli1808characterizing}.
		
		Certain features are specific to image datasets. Histogram of Oriented Gradients (HOG) \cite{dalal2005histograms} captures edge directions and intensities, while Local Binary Patterns (LBP) \cite{huang2011local} describe texture characteristics by encoding local intensity variations. Cho and Lee introduced two metrics in \cite{cho2021data} to assess image dataset quality: Msep measures class separability, while Mvar quantifies in-class variability.
		
		Decision tree-based meta-features assess dataset complexity by training a decision tree model on each dataset and extracting statistics, including the number of leaves, branches, nodes, and non-leaf nodes \cite{rivolli1808characterizing}. Additional features include the proportion of leaves per class, node-to-feature ratio, node-to-instance ratio, tree depth, probability of reaching a leaf randomly, and node importance scores.
		
		In \cite{leng2023complexity}, a new metric for measuring data classification complexity, called \( C^2M\_kNN \), is proposed. This method allows for better prediction of classification performance compared to other features that describe synthetic and 1D real datasets. The paper \cite{guan2020data} introduces the Distance-based Separability Index as an effective and model-independent measure of dataset separability, with potential applications in deep learning, data science, and AI interpretability. This metric is also evaluated on synthetic datasets and the CIFAR-10/100 dataset. In paper \cite{maillo2020redundancy}, two new metrics, Neighborhood Density (ND) and Decision Tree Progression (DTP), are proposed for assessing redundancy, complexity, and density in big data classification. These metrics help identify unnecessary data in large datasets, allowing for efficient preprocessing. The metrics, implemented in a Spark-based package, aid in reducing dataset size without significantly affecting accuracy, promoting smart data usage.
		\subsection{Decision-Making}
		Meta-features are widely used to guide dataset-level decisions such as classifier selection. For example, \cite{garcia2018classifier} leverages dataset complexity metrics for 1D classifier selection, while \cite{pereira2021evaluating} demonstrates that dimensionality reduction on meta-features accelerates training without sacrificing predictive performance. For spam detection, \cite{mekouar2021classifiers} applies an Analytic Hierarchy Process to rank and recommend top-$K$ classifiers across multiple metrics.
		
		Meta-learning also optimizes visual tasks. Frameworks like MetaDelta \cite{chen2021metadelta} automate few-shot classifier selection, \cite{hendryx2019meta} introduces meta-learning initialization for weakly supervised segmentation, and SSM-SAM \cite{leng2024self} integrates adaptive attention for medical imaging. As surveyed in \cite{luo2022meta}, classifier efficacy depends directly on dataset characteristics. Motivated by this, our framework uses meta-learning to map dataset descriptors to empirical performance, enabling efficient, data-driven classifier selection.
		
		However, automating decisions for image datasets presents distinct challenges. High-dimensional pixel data increases computational costs and overfitting risks. Extracting features demands either labor-intensive engineering or resource-intensive deep architectures that require massive labeled datasets. Real-world variability—such as noise, lighting shifts, and class imbalance—compounds these issues. Finally, the opaque nature of modern deep models turns model selection into a difficult trade-off between accuracy, latency, and interpretability, particularly in high-stakes domains like medical imaging.
		
		Due to these challenges, limited research has focused on quantifying dataset complexity and classifier selection for image datasets. While we cannot address all challenges comprehensively, our work aims to mitigate several key issues through a meta-learning framework based on dataset complexity measures. Our approach: (1) generates a meta-dataset capturing relevant dataset characteristics, and (2) trains a meta-model to predict classifier performance, thereby reducing—though not eliminating—the challenges in model selection. This partial solution focuses particularly on improving generalization through better understanding of dataset properties. For implementation, we categorize meta-features by their methodological perspectives and apply feature selection to identify the most impactful features for classifier prediction. The key contributions of this paper are as follows:
		
		\begin{itemize}
			\item {Meta-Learning Framework for Image Classification:} We introduce a meta-learning approach that predicts the best ML classifier for image datasets based on their inherent complexity measures.
			
			\item {Generation of Meta-Dataset:} We generate a meta-dataset that captures the characteristics and complexities of different image datasets. This meta-dataset is used to train a meta-model to predict classifier accuracy for new datasets.
			
			\item {Enhanced Classifier Performance and Generalization:} By leveraging the complexity measures and meta-learning approach, we improve the performance and generalization of machine learning systems, especially in image classification tasks.
			
			\item {Categorization and Selection of Meta-Features:} We categorize the various meta-features based on their methods and perspectives, ensuring a structured approach for the subsequent application of feature selection techniques.
			
			\item {Application of Feature Selection Methods:} We apply different feature selection techniques on the categorized meta-features to identify the most relevant ones for predicting classifier performance.
		\end{itemize}
		
		The remaining structure of the paper is as follows: In Section 2, the formulation of our work is presented. In Section 3, the details of the proposed framework are described. Section 4 provides an analysis of the results, and the final section concludes the paper.
		\section{Formulation of Decision-Making as a Machine Learning Problem}
		
		The origins of the algorithm selection problem trace back to a fundamental question in computational science: Why does a specific algorithm outperform others on certain problem instances? This inquiry, which lies at the core of optimization and machine learning research, was first formally modeled by John Rice in 1976. The significance of this framework is that it transitioned algorithm selection from a heuristic-based practice into a quantifiable scientific problem characterized by four fundamental components\cite{rice1976algorithm}.
		
		Building upon this foundational theory, the process of selecting an appropriate classifier for a given image dataset can be formulated as a meta-learning problem. In this context, the goal is to predict the optimal classifier by mapping the intrinsic characteristics of a dataset to the performance of various algorithms. This formulation typically involves three functional stages: feature extraction (characterizing the dataset), feature selection (identifying the most predictive metadata), and the training of a meta-model to automate the decision-making process.
		\subsection{Preprocessing}
		Given that images across different datasets often possess varying dimensions and aspect ratios, the initial step involves a preprocessing phase to ensure data uniformity. In this stage, all images are first resized to a fixed resolution to produce a standardized set of images, defined as:
		\[
		D^{\text{res}} = \{ \tilde{I}_1, \tilde{I}_2, \dots, \tilde{I}_N \}.
		\]
		
		Subsequently, depending on the specific type of meta-features required, certain operations are performed directly on the resized images, while others necessitate the transformation of images into feature vectors. To this end, a vectorization and dimensionality reduction mapping is defined:
		\[
		\phi : \tilde{I}_i \mapsto v_i \in \mathbb{R}^d
		\]
		Consequently, the resulting set of feature vectors is obtained as:
		\[
		D^{\text{vec}} = \{ v_1, v_2, \dots, v_N \}.
		\]
		\subsection{Feature Extraction and Meta-Dataset Creation}
		In this stage, the set of meta-features is extracted from the preprocessed data ($D^{\text{res}}$ or $D^{\text{vec}}$). These meta-features encompass metrics such as class separability, feature redundancy, intrinsic dimensionality, and various statistical indices that describe the underlying structure and inherent complexity of the data:
		\[
		X_{\text{meta}}(D^{\text{res or vec}}) = \{ f_1, f_2, \dots, f_n \}
		\]
		where each $f_i$ represents a meta-feature computed from either $D^{\text{res}}$ or $D^{\text{vec}}$.
		\subsection{Feature Selection}
		Once the meta-features are extracted, we apply feature selection methods to identify the most relevant features for predicting classifier performance. Feature selection helps reduce the dimensionality of the meta-dataset, removing irrelevant or redundant features, and retaining only those that significantly influence the model selection process. Common feature selection techniques include methods based on correlation, importance scores from regression models, and recursive elimination techniques. The goal is to retain a subset of meta-features \( X_{\text{meta}}^* \subset X_{\text{meta}} \) that are most predictive of classifier performance.
		
		Let \( X_{\text{meta}}^*(D^{\text{res or vec}}) \) represent the selected meta-features after applying the feature selection process:
		
		\[
		X_{\text{meta}}^*(D^{\text{res or vec}}) = \{ f_1^*, f_2^*, \dots, f_m^* \}
		\]
		where \( f_1^*, f_2^*, \dots, f_m^* \) are the selected meta-features after feature selection.
		
		\subsection{Training a Meta-Model}
		Once the meta-dataset is created and the feature selection step is completed, we train a meta-model \( M_{\text{meta}} \) that learns the relationship between the selected meta-features \( X_{\text{meta}}^*(D^{\text{res or vec}}) \) and the performance of various classifiers on the dataset \( D \). The goal is to predict the accuracy \( A_j \) of classifier \( j \) on the dataset \( D \), given the selected meta-features \( X_{\text{meta}}^*(D^{\text{res or vec}}) \). This transforms the problem into a regression problem where the meta-model is trained to predict classifier performance based on the dataset's meta-features.
		
		The meta-model \( M_{\text{meta}} \) can be written as:
		
		\[
		A_j = M_{\text{meta}}(X_{\text{meta}}^*(X_{\text{meta}}(D^{\text{res or vec}})))
		\]
		where
		\( A_j \) is the predicted accuracy of classifier \( j \) for dataset \( D \).
		\( M_{\text{meta}} \) is the trained machine learning model (e.g., regression, decision tree, neural network) that predicts classifier performance.
		
		\subsection{Select Proper Classifier}
		Once the meta-model is trained, it can be used to predict the proper classifier for any new, unseen dataset \( D' \). The decision-making process is now reduced to a machine learning task where input is the meta-features \( X_{\text{meta}}^*(D') \) of the new dataset \( D' \)
		and output is the predicted accuracy for each classifier \( A_j \), and the classifier with the highest predicted accuracy is chosen as the best classifier for dataset \( D' \). The decision rule can be expressed as:
		
		\[
		j^* = \arg\max_j A_j
		\]
		
		where \( j^* \) is the index of the classifier with the highest predicted accuracy \( A_j \).
		
		Thus, the classifier selection process is formalized as an optimization problem in which the goal is to maximize the predicted accuracy using the meta-model \( M_{\text{meta}} \). This turns the decision-making task into an end-to-end machine learning problem where both the model selection and accuracy prediction are based on dataset characteristics and the selected meta-features.
		\section{Proposed Method}
		Selecting an appropriate classifier for an image dataset is a crucial step in machine learning, as different classifiers exhibit varying levels of performance depending on the dataset characteristics. Image datasets often differ in terms of feature complexity, class distribution, resolution, and noise levels, making classifier selection a non-trivial task. Traditional approaches rely on empirical evaluation, where multiple classifiers are trained and tested on the dataset to determine the most suitable model. However, this process is computationally expensive, especially for large-scale image datasets.
		
		To address this challenge, we propose a framework that leverages dataset-specific features from different view points to predict proper classifier based on its probability or accuracy without the need for extensive training. By extracting statistical, structural, and complexity-based features from image datasets, we build a predictive model that estimates the performance of various classifiers, facilitating efficient model selection. Our approach aims to solve the problem while primarily providing insights into the relationship between dataset characteristics and classifier performance, thereby contributing to a more interpretable and automated machine learning workflow. 
		
\begin{figure*}[pb]
			\centerline{\includegraphics[width=14cm]{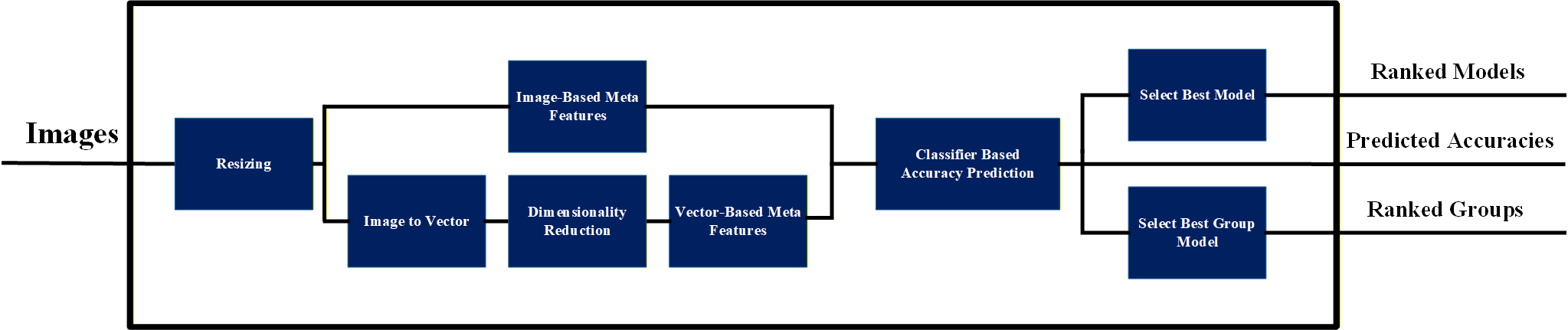}}
			\caption{The workflow of our approach.}
			\label{f1}
		\end{figure*}

		The workflow of our approach is illustrated in Figure \ref{f1}, encompassing several key contributions. Firstly, a diverse set of meta-features is extracted from each dataset, considering both image-based and vector-based characteristics. To transform images into one-dimensional representations, feature extraction techniques such as VGG19 \cite{simonyan2014very} and autoencoder are employed. Additionally, dimensionality reduction methods like Principal Component Analysis (PCA) \cite{jolliffe2016principal} and t-distributed Stochastic Neighbor Embedding (t-SNE) \cite{van2008visualizing} are utilized to minimize computational costs while preserving essential information. Next, classifiers are trained using the extracted one-dimensional data, and their accuracy is recorded. Based on the extracted meta-features and corresponding classifier performance, we generate a new set of descriptors for each dataset. These descriptors serve as a predictive dataset, enabling the selection of the most suitable classifier without the need for an exhaustive search.
		Our approach consists of two main parts: generating a dataset to train the trainable component in the workflow, and selecting the appropriate classifier for each dataset. These two parts involve several components, many of which are shared between them. However, the accuracy calculation and feature selection components are used only in the first part. In the following subsections, we provide a detailed explanation of all the components involved in these two parts.
		\subsection{Resizing}
		Since we work with datasets that have different applications and concepts, the images come in various dimensions. This variability could lead to an unfair comparison, so to make the analysis more consistent, we resize all the images to a similar dimension.
		
		\subsection{Convert Image to Vector}
		Two widely used approaches exist for converting an image into a vector representation: (i) directly converting the image into one-dimensional data without changing the size of data and (ii) extracting learned features using pre-trained models.
		
		The first approach involves flattening the image into a one-dimensional vector, which is simple and requires no special techniques. However, while easy to implement, flattening does not capture higher-level features or spatial relationships within the image. The second approach utilizes pre-trained neural network models such as autoencoders and deep classifiers like VGG19 to extract meaningful features. These models can be used in their pre-trained state or fine-tuned on our datasets. To evaluate their effectiveness, we explore both methods in our work.
		
		Flattening is advantageous because it is simple, computationally efficient, and universally applicable to any dataset. Unlike feature extraction using neural networks, flattening does not require a training phase; it simply reshapes the image into a one-dimensional vector. This makes it a quick and easy method for converting images into numerical representations. However, a key drawback is that flattening discards spatial relationships and hierarchical structures within the image.
		
		On the other hand, feature extraction using neural networks captures high-level patterns, textures, and semantic information that are crucial for classification. However, training deep networks from scratch is computationally expensive and requires large datasets to learn meaningful representations. This overhead can be mitigated by using pre-trained models, such as VGG19 or ResNet, which have already learned useful features from massive datasets like ImageNet. These models can directly extract robust and transferable features, significantly reducing the need for training.
		
		Despite their advantages, pre-trained models have limitations—they may not generalize well to domain-specific datasets, such as medical images or satellite imagery, where the distribution of features differs significantly from natural images in ImageNet. For example, a model trained on everyday objects may struggle to recognize microscopic cell structures in histopathology images. To overcome this, transfer learning is applied, where a pre-trained model is fine-tuned on a smaller dataset specific to the target domain. This approach allows the model to adapt its learned representations to new tasks while leveraging prior knowledge, improving performance in specialized applications \cite{weiss2016survey}.
		\subsection{Dimensionality Reduction}
		
		The use of dimensionality reduction comes after we convert the images into vectors. No matter the method used for this conversion, the resulting feature representations tend to be high-dimensional, which can slow down processing and increase memory usage. To handle this, we use PCA and t-SNE to reduce the dimensions, ensuring that the features are both more efficient computationally and still capture the key characteristics of the datasets.
		
		PCA is a linear dimensionality reduction technique that transforms the original high-dimensional data into a lower-dimensional space by projecting it onto a set of orthogonal axes called principal components. These components capture the directions of maximum variance in the data, allowing for a more compact representation while preserving important statistical information. PCA is particularly effective when dealing with highly correlated features, as it removes redundancy and highlights dominant patterns. However, since PCA assumes linear relationships, it may not be optimal for datasets with complex, nonlinear structures \cite{jolliffe2016principal}.
		
		Unlike PCA, t-SNE is a nonlinear dimensionality reduction technique that focuses on preserving the local structure of the data. It maps high-dimensional data into a lower-dimensional space while maintaining relationships between similar data points. However, it is computationally intensive and less suited for large-scale feature extraction compared to PCA. Additionally, since t-SNE is primarily used for visualization, it may not be ideal for downstream tasks that require consistent feature representations \cite{van2008visualizing}.
		\subsection{Accuracy Calculation}
		To create the dataset for our framework, we need an (input, output) pair for each sample. In earlier steps, the images in each dataset were converted into vector representations. Now, we define the target variable for each dataset sample. The goal of this work is to predict the classification accuracy of the most suitable classifier for a given dataset, helping us select the appropriate classifier. To achieve this, we evaluate multiple classifiers and calculate their respective accuracy scores for each dataset.
		
		For each dataset, we evaluate the performance of 15 different classifiers, representing a broad range of machine learning models. These classifiers are trained using the vectorized features of the datasets. After training, the accuracy scores from each classifier are used as target values for our predictive model. In other words, each dataset is paired with a set of accuracy scores, one for each classifier. These scores are then used to train a regression model that predicts classifier performance on unseen datasets and ranks the classifiers accordingly. This ranking can also serve as input for training a separate classifier that selects the most appropriate model, providing an alternative to the regression approach.
		\subsection{Meta-Features}
		
\begin{figure*}[pt]
			\centerline{\includegraphics[width=15cm]{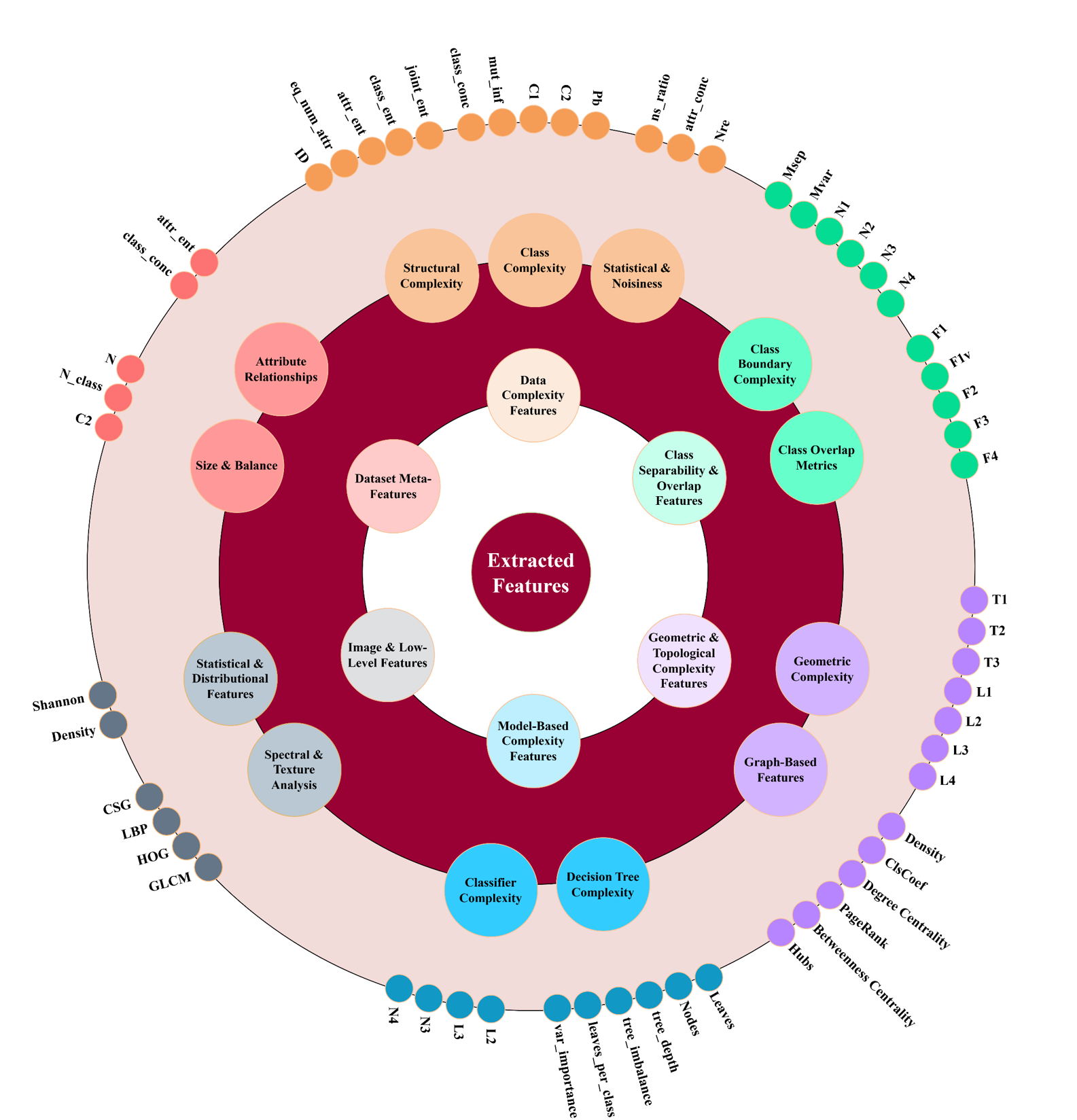}}
			\caption{The categories of used meta-features.}
			\label{f2}
		\end{figure*}

		Describing image datasets accurately is critical due to their inherent complexity and diversity. Extracting a single descriptor that fully captures all dataset characteristics is challenging. To address this, we extract a diverse set of meta-features that reflect various aspects of dataset complexity, including image structure, class separability, distributional properties, and graph-based characteristics. These features provide valuable insights that facilitate more accurate classifier selection. In Figure \ref{f2}, the whole meta-features and their categories are shown. Since there are lots of meta-features, they are organized into six main categories:
		
		\begin{itemize}
			\item {Data Complexity Features:} Capture intrinsic dataset properties such as intrinsic dimensionality, entropy measures, and class imbalance ratios.
			\item {Class Separability and Overlap Features:} Measure the degree of separability and overlap between classes to assess classification difficulty.
			\item {Geometric and Topological Features:} Characterize the geometric arrangement and topological structure of data points, including graph metrics like density and centrality.
			\item {Model-Based Complexity Features:} Derived from classifier models (e.g., decision trees) to quantify dataset complexity from a modeling perspective.
			\item {Image and Low-Level Features:} Include texture and spectral analysis such as Local Binary Patterns (LBP) and Histogram of Oriented Gradients (HOG) that capture image-specific characteristics.
			\item {Dataset-Level Meta-Features:} Describe overall dataset attributes, including sample size, number of classes, and attribute relationships.
		\end{itemize}
		
		This structured set of meta-features ensures a comprehensive representation of dataset properties, supporting more effective model selection.
		
		By leveraging this extensive set of meta-features, we provide a comprehensive characterization of datasets, ensuring more accurate classifier selection and enhancing the generalization of our regression model.
		\subsection{Feature Selection}
		Classifier selection can be approached in two ways: (1) predicting the accuracy of each classifier using a regression model and choosing the best-performing one, or (2) directly predicting the optimal classifier label. Since both methods rely on training data, the quality and relevance of the data are critical. To improve model performance and reduce computational overhead, it is essential to use features that effectively reflect the complexity of the dataset. Given the significance of feature quality, various selection methods have been proposed. In this study, we evaluate three such approaches: (i) correlation analysis between features and classifier accuracy, (ii) game theory-based feature selection, and (iii) Recursive Feature Elimination with Cross-Validation (RFECV).
		
		In the first approach, the correlation coefficient between each feature and classifier accuracy is calculated. We employ three correlation methods: Pearson, Spearman, and Kendall.  
		\begin{itemize}
			\item {Pearson correlation} examines the linear relationship between two continuous variables.
			\item {Spearman correlation} measures the strength and direction of a relationship using ranked values instead of raw data, making it a non-parametric measure that does not assume a specific data distribution.
			\item {Kendall correlation} evaluates relationships by counting the number of concordant and discordant observation pairs. Two observations are concordant if both variables increase or decrease together and discordant if one increases while the other decreases.
		\end{itemize}
		Since each method captures different aspects of correlation, we compute all three and select the highest correlation value for each feature. The correlation coefficient lies within the range [-1, 1]. For feature selection, the minimum and maximum correlation values for each feature are first calculated. Features with an absolute correlation value exceeding 0.8 are then selected from this set, resulting in 38 among 58 features. A high correlation suggests these features have greater predictive power, as they more effectively capture variations in classifier accuracy.
		
		The other two approaches, game theory-based feature selection and RFECV, are based on the trained model. In these two approaches, the value of each feature is calculated based on its effect on the training process, but from two different perspectives. For the game theory-based approach, the SHapley Additive exPlanations (SHAP) is used. SHAP explains the output of any machine learning model based on game-theory concepts, such as the Shapley value \cite{NIPS2017_7062}. The RFECV approach, on the other hand, is a recursive feature elimination algorithm that iteratively removes the least significant features using cross-validation.
		
		Both approaches require training a model first, after which feature importance scores are extracted. Since the goal is classifier accuracy prediction, we use regression models. The regression models are trained on the new dataset, where the extracted dataset features serve as inputs and each classifier’s accuracy is the target variable. After training, the importance of each feature is computed for every classifier separately. Since regression models can be linear, nonlinear, parametric, or non-parametric, we include a diverse set of models from each category, listed in Table \ref{tab1}. The three best models from this set is then selected to calculate the importance score.
		\begin{center}
			\begin{table*} [tp]
				\centering
				\caption{List of regression models used for predicting dataset-specific classifier accuracy.}
				\label{tab1}
				\resizebox{\textwidth}{!}{ 
					\begin{tabular}{|c|c|}
						\hline \multicolumn{2}{|c|}{Regression Models} \\
						\hline CatBoost Regressor (CatBoost) \cite{Prokhorenkova2018} & Extreme Gradient Boosting (xgboost) \cite{Chen2016} \\
						\hline Gradient Boosting Regressor (gbr) \cite{Zemel2000} & Random Forest Regressor (rf) \cite{Breiman2001} \\
						\hline Light Gradient Boosting Machine (LightGBM) \cite{Ke2017} & Ridge Regression (ridge) \cite{Hoerl2000} \\
						\hline Linear Regression (lr) \cite{Seber2003} & Elastic Net (en) \cite{Zou2005} \\
						\hline Lasso Regression (lasso) \cite{Tibshirani1996} & AdaBoost Regressor (ada) \cite{Ridgeway1999} \\
						\hline Huber Regressor (huber) \cite{Sun2020} & Bayesian Ridge (BR) \cite{Kalatzis2008} \\
						\hline Decision Tree Regressor (DT) \cite{Xu2005} & Orthogonal Matching Pursuit (omp) \cite{Pati1993} \\
						\hline Passive Aggressive Regressor (par) \cite{Crammer2006} & K Neighbors Regressor (knn) \cite{Kramer2011} \\
						\hline Lasso Least Angle Regression (llar) \cite{Fraley2009} & Dummy Regressor (dummy) \cite{Suits1957} \\
						\hline \multicolumn{2}{|c|}{Extra Trees Regressor (et) \cite{Geurts2006}} \\
						\hline	
					\end{tabular}
				}
			\end{table*}
		\end{center}
		After calculating the score for each feature in each classifier, the top features must be selected. In the first phase, 38 top features (equal to the number of features obtained from the correlation methods) are selected per classifier. Since there are 15 classifiers and three top regressors, the total number of selected features is 1620. These features are sorted in descending order based on their importance, and a weight is assigned to each feature according to its ranking, as follows:  
		\begin{itemize}
			\item The top-ranked feature receives a weight of 1.
			\item The least important feature receives a weight of \( \frac{1}{38} \).
			\item The remaining features are assigned weights in steps of \( \frac{1}{38} \).
		\end{itemize}
		Since a feature may appear across multiple classifiers, we compute the sum of its weights across all classifiers and assign it a final cumulative weight. Finally, the top 38 features with the highest total weights are selected.
		\subsection{Classifier Based Accuracy Prediction}
		To predict the accuracy of classifier models, two strategies can be employed: using a single model to estimate the accuracy of any given classifier based on its input features, or training a dedicated regression model for each classifier. Given that each classifier employs a distinct learning mechanism, the latter approach is adopted in this study. Various regression models, listed in Table \ref{tab1}, are trained using the selected features as input and the corresponding classifier accuracy as the target variable. Separate regressions are trained for each classifier, and the best-performing model is selected for final use. The performance of the regression models is evaluated using the MAE metric.
		\subsection{Select Best Model}
		Using the outputs from the previous step, the best-performing model can be identified. The classifier with the highest predicted accuracy is selected as the optimal choice. Additionally, a ranking of all classifiers based on their predicted accuracies can be generated as part of the output.
		\subsection{Select Best Group Model}
		Although each classifier has its own learning mechanism, they may produce similar results when applied to a given dataset. When classifiers yield comparable performance, they can be assigned the same rank for that dataset. To identify groups of classifiers exhibiting similar behavior, clustering is employed. The optimal number of clusters is determined using elbow method. Once the clusters are formed, classifiers are grouped accordingly, and instead of ranking individual classifiers as in the previous step, the ranks of the identified groups are used.
		
		The output of the proposed framework includes the predicted accuracies of the classifiers, their individual rankings, and the rankings of classifier groups.
		\section{Experimental Results}
		The proposed framework consists of several components, each with different selection methods for its respective part. In this section, we provide detailed explanations of the implementation, the results for each part, and the analysis of those results.
		\subsection{Dataset}
		Our experimental setup utilizes 56 image datasets spanning multiple domains, such as medicine, nature, and industry. We randomly allocated 28\% (n=16) of these datasets for testing, reserving the remaining 40 for training. This selection covers a broad spectrum of semantic concepts and structural characteristics (e.g., varying sample sizes and class counts) to facilitate a robust exploration of the problem space. For computational uniformity, all images were resized to a fixed resolution of $96 \times 96$ pixels.
		\subsection{Combination of Flatten and Dimensionality Reduction}
		To convert images into vectors and flatten them, we use a simple flattening method along with two neural network models. Since the image sizes in some datasets are large and vary, the simple flattening method is not sufficient for our approach due to the time-consuming meta-feature extraction process. Therefore, two neural network models are employed in this work. Any classification or neural network model capable of extracting features from images could be used, but the models chosen here are VGG19 and AutoEncoder. For each model, two scenarios—pre-trained and transfer learning—can be applied. To evaluate both options, the pre-trained model is used for VGG19, while for AutoEncoder, the pre-trained model is applied first, followed by training for 100 epochs on each dataset. In this process, each image in the dataset is converted into a feature vector that captures the image's properties.
		
		The feature vectors extracted by these two models have dimensions of 100 and 288, which are still large for effective meta-feature extraction. Therefore, two dimensionality reduction techniques—PCA and t-SNE—are applied. For PCA, the dimensionality is reduced such that the reconstructed data retains 90\% of the original variance. For t-SNE, the features are reduced to a fixed dimension of 10. As a result, we obtain four sets of vector representations for each image.
		
		To compare these four sets, the correlation between the meta-features extracted from each set and the accuracy of the classifiers is calculated. Three methods—Pearson, Spearman, and Kendall—are used to calculate the correlation, and the number of features which have correlation greater than 0.8 is computed for each set. The results show that features extracted using an AutoEncoder with transfer learning, combined with PCA for dimensionality reduction, exhibit a stronger correlation with classifier performance. These results are presented in Table \ref{tab2}.
		\begin{center}
			\begin{table*} [thp]
				\centering
				\caption{Number of features with absolute value correlation greater than 0.8 for each feature set.}
				\label{tab2}
				\begin{tabular}{|c|c|c|c|c|}
					\hline  Features&AE\_PCA&VGG19\_PCA&AE\_t-SNE&VGG19\_t-SNE\\
					\hline Number&184&154&115&127\\
					\hline	
				\end{tabular}
			\end{table*}
		\end{center}
		\subsection{Compare Feature Selection}
		Three different feature selection methods are employed in this study. For the training-based methods, a set of regression models is first trained for each classifier, after which the top three performing regressors are selected per classifier. The variation in selected regression models across classifiers supports the decision to use separate training processes for each. The three selected regression models for each classifier are summarized in Table \ref{tab4}.
		\begin{table}[h!]
			\centering
			\caption{Selected Regression Models for Each Classifier}
			\label{tab4}
			\begin{tabular}{|c|c|}
				\hline
				\textbf{Classifier} & \textbf{Selected Regression Models} \\
				\hline
				svm & et, gbr, rf \\
				knn & et, gbr, rf \\
				mpl3 & br, gbr, rf \\
				mpl5 & et, rf, gbr \\
				rf & ada, lightgbm, gbr \\
				dt & et, rf, gbr \\
				nb & et, rf, gbr \\
				et & dt, gbr, lightgbm \\
				gbc & et,gbr,rf \\
				qda & et, gbr, lightgbm \\
				lda &  et, gbr, lightgbm \\
				lr & et, lightgbm,gbr \\
				rc & et, gbr, lightgbm \\
				abc & et, gbr, lightgbm \\
				dummy & et,dt,gbr  \\
				\hline
			\end{tabular}
		\end{table}
		To evaluate the impact of these methods, the results are compared in this section. Since the correlation method directly uses the classifier's accuracy, the features selected by the last two methods (RFECV and SHAP) are compared with those selected by the correlation method. Table \ref{tab3} presents a comparison of the selected features across the three methods. In the second column, the blue rows highlight the features that are common between correlation and RFECV. In the third column, the yellow rows highlight the features that are common between correlation and SHAP. The results indicate:
		\begin{itemize}
			\item 26 features are shared between correlation Analysis and RFECV.  
			\item 21 features are shared between correlation Analysis and SHAP.  
			\item 12 features are common across all three methods.  
		\end{itemize}
		The significant overlap among the selected features indicates that the regression models tend to prioritize features that effectively capture trends in classifier accuracy. Interestingly, image-specific features such as texture and edge-related metrics were not among the top features selected based on correlation analysis. However, texture features were included among those selected by RFECV, and both texture and edge features were identified as important using SHAP analysis.
		\begin{center}
			\begin{table*} [thp]
				\small
				\caption{Comparison of features selected with three feature selection methods.}
				\label{tab3}
				\begin{tabular}{|c|c|c|}
					\hline
					\textbf{Correlation} & \textbf{Pycaret} & \textbf{SHAP} \\
					\hline
					\cellcolor{white}nre & \cellcolor{lightblue}nodes\_per\_inst & \cellcolor{white}attr\_conc \\
					\cellcolor{white}pb & \cellcolor{lightblue}sil & \cellcolor{lightyellow}n2 \\
					\cellcolor{white}sil & \cellcolor{lightblue}n1 & \cellcolor{lightyellow}closeness\_centrality \\
					\cellcolor{white}class\_conc & \cellcolor{lightblue}n3 & \cellcolor{lightyellow}class\_ent \\
					\cellcolor{white}class\_ent & \cellcolor{lightblue}n4 & \cellcolor{lightyellow}f3 \\
					\cellcolor{white}eq\_num\_attr & \cellcolor{lightblue}pb & \cellcolor{lightyellow}nre \\
					\cellcolor{white}joint\_ent & \cellcolor{lightblue}vdb & \cellcolor{lightyellow}degree\_centrality \\
					\cellcolor{white}leaves & \cellcolor{lightblue}density & \cellcolor{white}ch \\
					\cellcolor{white}leaves\_branch & \cellcolor{white}int & \cellcolor{lightyellow}class\_conc\\ 
					\cellcolor{white}leaves\_corrob & \cellcolor{lightblue}f3 & \cellcolor{white}mut\_inf \\
					\cellcolor{white}leaves\_homo & \cellcolor{white}l2 & \cellcolor{lightyellow}leaves\_per\_class \\
					\cellcolor{white}leaves\_per\_class & \cellcolor{lightblue}lsc & \cellcolor{lightyellow}joint\_ent\\
					\cellcolor{white}nodes & \cellcolor{white}l3 & \cellcolor{lightyellow}n\_class \\
					\cellcolor{white}nodes\_per\_attr & \cellcolor{white}f1v & \cellcolor{white}hog \\
					\cellcolor{white}nodes\_per\_inst & \cellcolor{lightblue}f4 & \cellcolor{lightyellow}sil \\
					\cellcolor{white}nodes\_per\_level & \cellcolor{lightblue}n2 & \cellcolor{white}lbp \\
					\cellcolor{white}nodes\_repeated & \cellcolor{lightblue}hubs & \cellcolor{lightyellow}eq\_num\_attr \\
					\cellcolor{white}tree\_depth & \cellcolor{lightblue}joint\_ent & \cellcolor{lightyellow}density \\
					\cellcolor{white}tree\_imbalance & \cellcolor{lightblue}nodes\_repeated & \cellcolor{white}attr\_ent \\
					\cellcolor{white}tree\_shape & \cellcolor{lightblue}nodes\_per\_level & \cellcolor{white}ns\_ratio \\
					\cellcolor{white}vdb & \cellcolor{lightblue}nodes\_per\_attr & \cellcolor{lightyellow}csg \\ 
					\cellcolor{white}f3 & \cellcolor{lightblue}leaves\_branch&\cellcolor{white}edge\_betweenness\_centrality \\ 
					\cellcolor{white}f4 & \cellcolor{lightblue}tree\_depth &\cellcolor{white}glcm \\
					\cellcolor{white}n1 & \cellcolor{lightblue}msep & \cellcolor{lightyellow}eigenvector\_centrality \\
					\cellcolor{white}n2 & \cellcolor{lightblue}class\_ent & \cellcolor{lightyellow}msep \\
					\cellcolor{white}n3 & \cellcolor{lightblue}nre & \cellcolor{white}vdu \\
					\cellcolor{white}n4 & \cellcolor{lightblue}tree\_shape & \cellcolor{lightyellow}f4 \\
					\cellcolor{white}t1 & \cellcolor{white}ch & \cellcolor{white}mvar \\
					\cellcolor{white}lsc & \cellcolor{white}lbp & \cellcolor{white}clscoef \\
					\cellcolor{white}density & \cellcolor{white}attr\_conc & \cellcolor{lightyellow}bp \\
					\cellcolor{white}hubs & \cellcolor{white}vdu & \cellcolor{lightyellow}leaves\_homo \\
					\cellcolor{white}n\_class & \cellcolor{lightblue}closeness\_centrality & \cellcolor{lightyellow}hubs \\
					\cellcolor{white}csg & \cellcolor{lightblue}class\_conc & \cellcolor{lightyellow}betweenness\_centrality \\
					\cellcolor{white}degree\_centrality & \cellcolor{white}f2 & \cellcolor{white}shannon \\
					\cellcolor{white}eigenvector\_centrality & \cellcolor{white}clustering & \cellcolor{white}f1 \\
					\cellcolor{white}closeness\_centrality & \cellcolor{white}l1 & \cellcolor{white}clustering \\
					\cellcolor{white}betweenness\_centrality & \cellcolor{white}var\_importance.mean & \cellcolor{white}pagerank \\
					\cellcolor{white}msep & \cellcolor{lightblue}nodes & \cellcolor{white}int \\
					\hline
				\end{tabular}
			\end{table*}
		\end{center}
		\subsection{Classifier-Based Accuracy Prediction}
		For training our regression models, the 38 features that are selected by the correlation method are used as input, and the accuracy of each classifier is used as the target variable. The 15 classifiers used in this work are Support Vector Machine (SVM), K-Nearest Neighbors (KNN), Multilayer Perceptron with 3 Layers (MLP3), Multilayer Perceptron with 5 Layers (MLP5), Random Forest (RF), Decision Tree (DT), Naive Bayes (NB), Extra Trees (ET), Gradient Boosting (GB), Quadratic Discriminant Analysis (QDA), Linear Discriminant Analysis (LDA), Logistic Regression (LR), Ridge, AdaBoost (AB), and Dummy. 
		
		Each classifier brings a distinct perspective to the training process. LR models class probabilities using a logistic function, while Ridge incorporates an $L_2$ penalty to reduce overfitting. SVM identifies the optimal separating hyperplane in high-dimensional settings, and LDA projects features along axes that maximize class separation. In contrast, QDA models each class as an independent Gaussian distribution, while NB applies Bayes' theorem under a conditional feature independence assumption. Instance- and rule-based methods include KNN, which assigns labels via local majority voting, and DT, which recursively partitions feature space for interpretability. Tree ensembles expand on this: RF aggregates diverse trees to mitigate variance, ET introduces randomized split thresholds to limit overfitting, GB sequentially minimizes residual errors for high predictive accuracy, and AB iteratively upweights difficult instances. Neural representations are captured by MLP3 and MLP5 to model patterns across three and five layers, respectively. Finally, Dummy serves as a heuristic baseline to benchmark empirical performance.
		
		Since the classifiers offer different perspectives, regression models in Table \ref{tab1} are trained separately for each classifier. For each classifier, the best regression model are selected. Prediction errors are computed on the test set, with the corresponding MSE and MAPE metrics reported in Table \ref{tab5}.
		\begin{table}[pht]
			\centering
			\caption{Regressor Performance Metrics (MAE, MSE, MAPE) for each Classifier}
			\label{tab5}
			\begin{tabular}{ccccc}
				\toprule
				\textbf{Classifier} &\textbf{Regressor} & \textbf{MAE} & \textbf{MSE} & \textbf{MAPE} \\
				\midrule
				svm&et & 0.0372 & 0.0025 & 0.0711 \\
				knn&et & 0.0241 & 0.0011 & 0.0409 \\
				mlp3&et & 0.0630 & 0.0077 & 0.1995 \\
				mlp5&et & 0.1331 & 0.0321 & 0.4662 \\
				rf&ada & 0.0002 & 0.0000 & 0.0002 \\
				dt&et & 0.0087 & 0.0003 & 0.0097 \\
				nb&et & 0.0508 & 0.0038 & 0.1220 \\
				et&dt & 0.0001 & 0.0000 & 0.0001 \\
				gbc&et & 0.0431 & 0.0036 & 0.0613 \\
				qda&gbr & 0.0669 & 0.0092 & 0.1378 \\
				lda&et & 0.0464 & 0.0033 & 0.1111 \\
				lr&et & 0.0521 & 0.0041 & 0.1776 \\
				rc&et & 0.0421 & 0.0028 & 0.1132 \\
				abc&et & 0.0501 & 0.0042 & 0.1580 \\
				dummy&dt & 0.0460 & 0.0068 & 0.2085 \\
				\bottomrule
			\end{tabular}
		\end{table}
		\subsection{Select Best Model}
		Based on the predicted accuracies, a ranking of classifiers can be generated for each dataset, with the highest rank corresponding to the best-performing model. The ranking accuracy achieved using all features is 50\%, while using correlation-based selected features yields 40\% accuracy. During the analysis, we observed that some classifiers exhibit similar or closely related behavior on certain datasets. Consequently, we propose a method for selecting the best-performing group of classifiers.
		\subsection{Select Best Group Model}
		In this section, we propose an alternative approach to ranking classifiers for a given dataset by combining clustering techniques with regression-based accuracy predictions. Instead of directly ranking classifiers by predicted accuracy, we cluster them based on their performance on the training data and use these clusters to refine the ranking process. This method addresses the observation that some classifiers demonstrate similar performance levels, where small differences in accuracy do not justify separate rankings in the selection process.
		
		\subsubsection*{Clustering Classifiers Based on Performance}
\begin{figure}[pt]
			\centerline{\includegraphics[width=9cm]{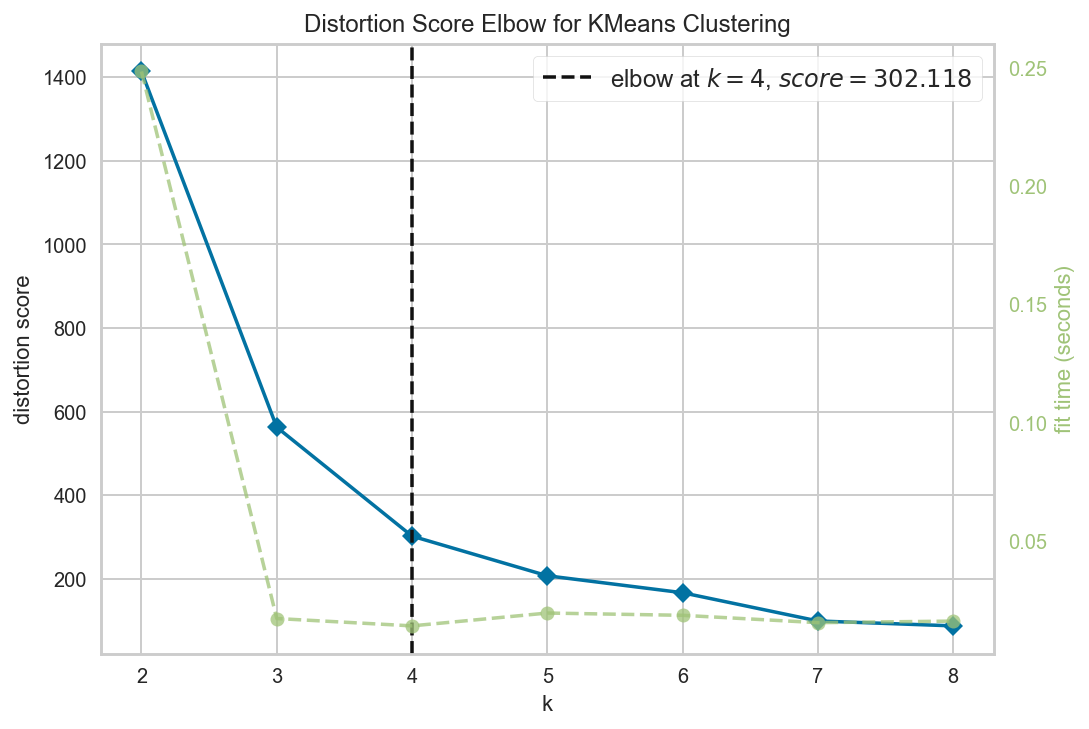}}
			\caption{Elbow output.}
			\label{f6}
		\end{figure}

		To cluster the classifiers, we use the actual accuracies of 15 classifiers on the training datasets. The K-means clustering algorithm is applied to group the classifiers based on their predicted accuracy values. The key parameter in clustering is the number of clusters, which is determined using the elbow method. Using this approach, four clusters are identified. The output of the elbow is shown in Figure \ref{f6}. The final four clusters are as follows:
		\begin{itemize}
			\item \textbf{Cluster 1:} RF, DT, ET, GB
			\item \textbf{Cluster 2:} SVM, KNN, MLP3, MLP5, QDA
			\item \textbf{Cluster 3:} NB, LDA, LR, RC, ABC
			\item \textbf{Cluster 4:} Dummy
		\end{itemize}
		The rationale behind clustering is based on the similarity in performance among certain classifiers. By comparing their predicted accuracies, we observe that the differences between classifiers within the same cluster are often minimal and do not significantly impact their relative effectiveness on a given dataset. Clustering simplifies the ranking process by grouping classifiers with comparable predictive capabilities, reducing the complexity of distinguishing between marginally different performers.
		
		To evaluate the effectiveness of this approach, we follow these steps: First, we compute the true ranking of classifiers on the test datasets based on their actual classification accuracy. Next, we predict the accuracy of each classifier on the test datasets using the regression models selected for each classifier (as shown in Table \ref{tab4}) and establish a predicted ranking based on these values. The predicted ranking represents the order in which classifiers are recommended for a given dataset.
		
		Clustering the classifiers before ranking prediction offers several advantages. First, it reduces the framework's sensitivity to small, often statistically insignificant accuracy differences, focusing on broader performance trends instead. Second, it improves interpretability by grouping classifiers with similar strengths, making it easier for practitioners to identify which families of methods are most suitable for a given dataset. For instance, Cluster 1 (RF, DT, ET, GB) represents tree-based ensemble methods known for their robustness and generalization, while Cluster 2 (SVM, KNN, QDA, MLP3, MLP5) includes methods sensitive to data geometry and separability.
		
		By applying this approach to the test datasets, we assess whether the predicted ranking, informed by both clustering and regression, captures the true order of classifier performance.
		
		The results of this evaluation, including the accuracy of the cluster-based ranking predictions, will be presented in the next section once the clustering-specific metrics are fully analyzed. This hybrid approach, combining clustering and regression, aims to balance precision and practicality in classifier selection, enhancing the framework’s utility for diverse image classification tasks.
		
		\subsection{Results}
		In this section, we present the results of the proposed clustering and regression-based framework for predicting the ranking of 15 classifiers on test datasets. The framework leverages K-means clustering to group classifiers into four performance-based clusters and employs regression models to predict classifier accuracies, which are then used to establish a predicted ranking. 
		
		The prediction accuracies for each of the 15 ranking levels are summarized in Table \ref{tab:results}. The framework achieves a mean prediction accuracy of 86.15\% across all ranks, demonstrating good performance in identifying the relative effectiveness of classifiers for a given dataset. 
		
		\begin{table*}[h]
			\centering
			\caption{Prediction Accuracy and Cumulative Average for Classifier Rankings Across 15 Levels}
			\label{tab:results}
			\begin{tabular}{ccc}
				\toprule
				\textbf{Rank} & \textbf{Prediction Accuracy (\%)} & \textbf{Cumulative Average (\%)} \\ \midrule
				1 & 95.3 & 95.30 \\ \addlinespace[1mm]
				2 & 100.0 & 97.65 \\ \addlinespace[1mm]
				3 & 98.4 & 97.90 \\ \addlinespace[1mm]
				4 & 81.3 & 93.75 \\ \addlinespace[1mm]
				5 & 75.0 & 90.00 \\ \addlinespace[1mm]
				6 & 81.3 & 88.55 \\ \addlinespace[1mm]
				7 & 90.6 & 88.84 \\ \addlinespace[1mm]
				8 & 78.1 & 87.50 \\ \addlinespace[1mm]
				9 & 78.1 & 86.46 \\ \addlinespace[1mm]
				10 & 71.9 & 85.00 \\ \addlinespace[1mm]
				11 & 89.1 & 85.37 \\ \addlinespace[1mm]
				12 & 87.5 & 85.55 \\ \addlinespace[1mm]
				13 & 85.9 & 85.58 \\ \addlinespace[1mm]
				14 & 89.1 & 85.83 \\ \addlinespace[1mm]
				15 & 90.6 & 86.15 \\ \midrule
			\end{tabular}
		\end{table*}
		
\begin{figure}[h]
			\centering
			\includegraphics[width=0.5\textwidth]{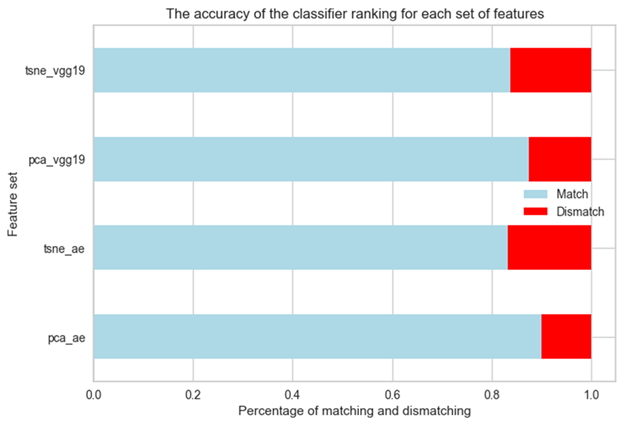}
			\caption{Number of matched and mismatched predictions for each feature extraction method (\texttt{pca\_ae}, \texttt{pca\_vgg19}, \texttt{tsne\_ae}, \texttt{tsne\_vgg19}). Blue bars represent correct predictions (matches), while red bars indicate incorrect predictions (mismatches).}
			\label{feature_extraction_matches_mismatches}
		\end{figure}

		We also evaluate the impact of different feature extraction methods on the framework’s performance. Four methods are compared: \texttt{pca\_ae}, \texttt{pca\_vgg19}, \texttt{tsne\_ae}, and \texttt{tsne\_vgg19}. Figure \ref{feature_extraction_matches_mismatches} shows the number of matches and mismatches for each method. Among these, feature extraction using \texttt{pca\_ae}—which combines an Autoencoder with PCA compression—yields the best performance in the proposed framework. Both correlation and accuracy metrics confirm the superior results of \texttt{pca\_ae}.
		
		This outcome indicates that multiple factors influence the results. First, the number of features extracted from the pre-trained models affects performance. Second, applying transfer learning alongside pre-training contributes to improved results. Third, the dimensionality reduction approach, tailored specifically to each dataset, also plays a critical role. For the Autoencoder, the first two factors are considered, while for PCA, instead of using a fixed dimension, we select the number of components based on a fixed reconstruction accuracy, resulting in a variable number of features for each dataset.
		The proposed clustering and regression-based framework demonstrates strong performance in predicting the rankings of 15 classifiers across various test datasets. The framework combines K-means clustering to group classifiers by performance similarity and regression models to estimate individual classifier accuracies, which are then used to derive predicted rankings. Key observations from Table \ref{tab:results}:
		
		High accuracy at top ranks:
		The prediction accuracy for the highest ranks (1st to 3rd) is exceptionally high, reaching above 95\%, with rank 2 achieving a perfect 100\% accuracy. This indicates that the framework is very effective at correctly identifying the top-performing classifiers for a dataset, which is often the most critical in practical applications.
		
		Gradual decline in mid-rank accuracy:
		From ranks 4 to 10, prediction accuracy shows a moderate decline, fluctuating between approximately 72\% and 90\%. This suggests that while the framework remains fairly accurate at identifying mid-tier classifier rankings, some ambiguity or overlap exists among classifiers ranked in this middle range, possibly due to closer performance levels.
		
		Stable accuracy for lower ranks:
		Interestingly, the accuracy for ranks 11 to 15 remains relatively stable and consistently high (around 85\% to 91\%). This implies that the framework is also reliable in distinguishing the lower-performing classifiers, helping to avoid recommending less effective models.
		
		\subsection{Comparison}
		To compare our work, we need to find articles that provide relevant features or datasets that can be used for comparison. A similar study that utilizes comparable features is \cite{garcia2018classifier}. In that work, complexity features are extracted from nearly 140 1D datasets, and the accuracy of four classifiers is calculated for each dataset. The goal of that study was to recommend the best classifier for each dataset based on these features and the classifiers' accuracies.
		
		However, since our datasets are image-based and the features we use differ from those in \cite{garcia2018classifier}, it is important to compare the results and assess the impact of our proposed features. To this end, we applied the 22 features from \cite{garcia2018classifier} to our dataset and evaluated their performance. Using these features, we achieved a ranking accuracy of 85.41\% on the test set. For a fair comparison, we also selected the first 22 features out of our 38 features, matching the same feature count. In this case, the ranking accuracy improved to 87.60\%. These findings indicate that our features, which are selected from different view point for image-based datasets, provide better performance than those used in \cite{garcia2018classifier}. Moreover, the improvement observed when using 22 features instead of all 38 suggests that a higher correlation threshold for feature selection may help retain only the most relevant features and further enhance performance.
		
		Instead of using the 38 selected features, we also conducted experiments with only the 12 features, which are common to the three selected feature methods. The results show that using these features reduces the complexity of feature extraction by nearly 70\%, the accuracy decreases 5.32\% (from 86.15\% to 81.56\%).
		
		Among the 38 selected features, some describe the generated tree after training a decision tree model. Since DT is one of our classifiers, we remove these features (13 features) to avoid any influence on our decisions. As a result, the accuracy decreases slightly from 86.15\% to 85.0\%.
		
		\section{Comparison with Related Work}
		
		Several recent studies have examined the challenges of image classification under diverse conditions, including industrial environments, noisy data, and cross-dataset variability. While these works provide valuable insights into classifier behavior, their objectives and methodological focus differ substantially from our proposed meta-learning framework for classifier recommendation.
		
		The work of \cite{Sepulveda2019IndustrialImages} investigates image classification in industrial inspection settings, where datasets are typically domain-specific and collected under controlled acquisition conditions. Their primary objective is to optimize classification performance within a fixed industrial application domain. The study focuses on improving predictive accuracy in a specialized context rather than addressing cross-domain generalization or automated model selection.
		
		Similarly, \cite{DeHoog2024NoisyImages} examine classifier robustness under noisy image conditions. Their analysis evaluates how classification performance degrades as noise levels increase, providing insights into model reliability in adverse or corrupted data scenarios. The emphasis is placed on robustness evaluation through controlled perturbations, offering a deeper understanding of classifier stability under noise.
		
		The study by \cite{theriault2025} highlights empirical performance variation of classifiers across multiple datasets, reinforcing the implications of the No Free Lunch theorem. Their findings demonstrate that no single classifier consistently outperforms others across all image distributions. The work primarily provides an empirical analysis of cross-dataset variability and underscores the dependency of classifier performance on dataset characteristics.
		
		In a different domain, \cite{eberlein2024} investigate the effect of data complexity on classifier performance in the context of software defect prediction, relying on tabular software engineering datasets rather than image data. Their study analyzes how complexity measures influence classification outcomes and shows that classifier effectiveness strongly depends on dataset properties. This further supports the idea that no universally optimal classifier exists across all problems.
		
		In contrast to these studies, our work proposes a unified and scalable meta-learning framework specifically designed for heterogeneous image datasets. Rather than optimizing performance within a single domain, analyzing robustness under specific perturbations, or solely conducting empirical performance comparisons, we aim to predict classifier performance directly from dataset characteristics. By extracting comprehensive image-based meta-features using autoencoders, pre-trained networks, and dimensionality reduction techniques, we train regression models to estimate classifier accuracies without exhaustive retraining. Furthermore, we incorporate clustering methods to group classifiers with similar performance patterns, simplifying the recommendation process.
		
		\section{Conclusion}
		In this paper, we proposed a meta-learning framework designed to predict the performance of various classifiers on image datasets by leveraging an extensive set of dataset complexity measures (meta-features). By extracting, selecting, and utilizing these meta-features, our approach enables efficient and accurate prediction of classifier accuracies without the need for exhaustive training and evaluation of all candidate models.
		
		The framework employs feature extraction techniques such as autoencoders and pre-trained deep networks, combined with dimensionality reduction methods like PCA and t-SNE, to represent image datasets effectively. Through rigorous feature selection and the training of dedicated regression models for each classifier, the system identifies the most relevant features that influence classifier performance. Furthermore, clustering classifiers based on performance similarities simplifies decision-making by grouping models with comparable accuracies, improving interpretability and robustness.
		
		Experimental results on a diverse set of 56 image datasets demonstrate that our method achieves a high average accuracy in predicting classifier rankings, notably excelling in identifying top-performing models. The approach also highlights the importance of using image-specific features and tailored meta-features to improve prediction performance.
		
		Overall, this work contributes to the field by providing a scalable, interpretable, and effective method for classifier recommendation in image classification tasks. Future research may extend this framework to other types of data and explore more advanced meta-features and meta-models to further enhance prediction accuracy and decision-making capabilities.

		
		
		
		
		\bibliographystyle{elsarticle-num} 
		\bibliography{refs.bib}
	\end{document}